\documentclass[twoside]{article}

\usepackage[accepted]{aistats2022}
\usepackage{amsthm}
\usepackage{amssymb}

\usepackage{algorithm}
\usepackage{algorithmic}
\def\calH{\mathcal{H}}
\def\bw{\mathbf{w}}

\usepackage{lipsum}
\usepackage{graphicx}
\newtheorem{lemma}{Lemma}
\usepackage{hyperref}

\begin{document}

%

%

\twocolumn[

\aistatstitle{Improving Attribution Methods by Learning Submodular Functions}

\aistatsauthor{Piyushi Manupriya\And Tarun Ram Menta\And  J. Saketha Nath\And Vineeth N Balasubramanian}

\aistatsaddress{ Indian Institute of Technology, Hyderabad } ]

\begin{abstract}
  This work explores the novel idea of learning a submodular scoring function to improve the specificity/selectivity of existing feature attribution methods. Submodular scores are natural for attribution as they are known to accurately model the principle of diminishing returns. A new formulation for learning a deep submodular set function that is consistent with the real-valued attribution maps obtained by existing attribution methods is proposed. The final attribution value of a feature is then defined as the marginal gain in the induced submodular score of the feature in the context of other highly attributed features, thus decreasing the attribution of redundant yet discriminatory features. Experiments on multiple datasets illustrate that the proposed attribution method achieves higher specificity along with good discriminative power. The implementation of our method is publicly available at \href{https://github.com/Piyushi-0/SEA-NN}{https://github.com/Piyushi-0/SEA-NN}.
\end{abstract}

\section{INTRODUCTION}
\vspace{-8pt}
Deep neural networks (DNNs) have shown state-of-the-art performance in diverse application domains, including complex tasks such as image recognition, video synthesis, speech-to-text conversion and autonomous navigation, to name a few. 
While advancements in deep learning have led to increasing accuracy scores, this has often been at the expense of the interpretability of the neural network's decisions. 
Our work focuses on attribution algorithms for interpreting neural networks. These algorithms output an attribution score heatmap that represents the feature-wise contribution of each input feature towards the prediction, a topic that has attracted significant interest in recent years \cite{DBLP:journals/corr/ZeilerF13, gbp, selvarajugrad,gradcamp, shrikumar2017learning, lundberg2017neurips, pmlr-v70-sundararajan17a, pmlr-v97-chattopadhyay19a}. 
\begin{figure}[t]
\label{motiv-ex}
    \centering
    \includegraphics[width = \columnwidth]{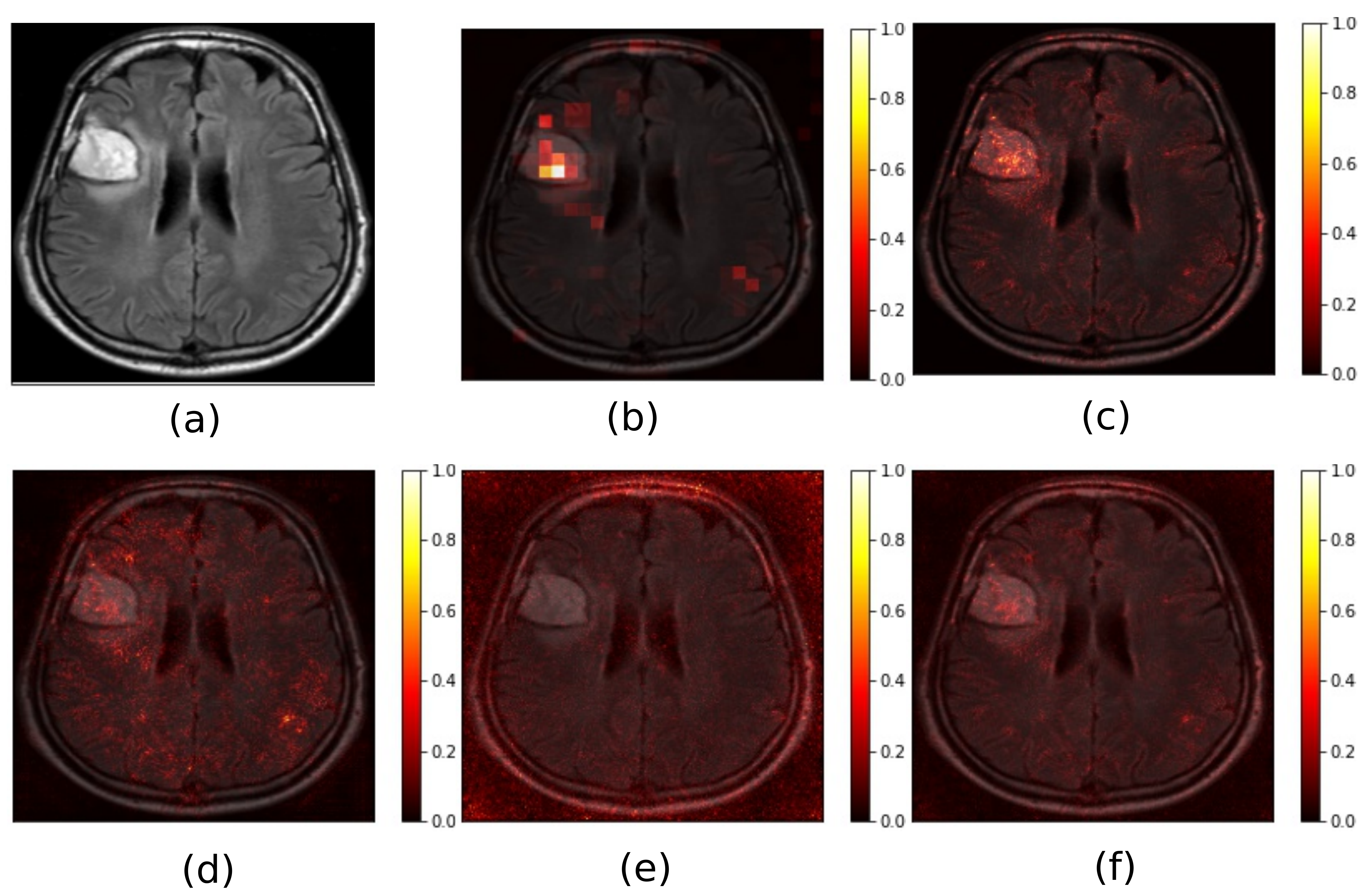}
    \vspace{-15pt}
    \caption{Example image and attribution maps overlaid on the image from the Brain Tumor Dataset: (a) Image with brain tumor; (b) SEA-NN (proposed) map; (c) Integrated Gradient map; (d) Vanilla Gradient map; (e) Smooth Integrated Gradient map; (f) Agg-Mean map of (c)-(e) maps. More such examples are provided in our results section.}
    \label{fig:dsf-inputs}
\end{figure}

As shown in Fig \ref{motiv-ex}, attribution algorithms provide a normalized score for each input feature (pixels in case of image-based problems, where such methods are extensively used), which denotes its contribution to the prediction on the given input. Our approach in this work is based on the observation that attribution heatmaps obtained by existing methods are not sharply focused around the discriminatory pixels and are often spread out across the image (see Fig~\ref{motiv-ex}). In other words, though existing algorithms achieve good sensitivity, they do not seem to achieve high specificity/selectivity. In applications like brain tumor detection where the region of interest usually lies in a very small area, specificity can be especially critical. The key idea in our work is to employ a monotone submodular set function to score -- and subsequently ensemble -- known heatmaps, such that the scores for the most specific of the given maps saturate (i.e. adding any more pixels to the map will not provide benefit).

However, such a submodular function is neither known nor designed in existing literature. We hence propose a novel algorithm that \textit{learns} one based on the heatmaps with existing attribution algorithms. Learning such a submodular function across multiple attribution algorithms provides us a mechanism to combine these methods, as well as provides attribution maps that are consistent across methods. The works closest to ours include \cite{rieger}, which performs pixel-wise averaging of attribution maps, and \cite{NIPS2017_6993}, which uses a notion of weak submodularity to select top-$k$ important pixels in a streaming setting. We compare against \cite{rieger} in this work, and show that our ensembled attribution map guided by a learned submodular scoring function outperforms simple linear averaging. ~\cite{NIPS2017_6993} however focuses on interpretability in a streaming setting, which is different from our objective. Besides, they focus on selection of a discrete subset of features, while we focus on scoring the heatmaps.  Also, as noted by them, the function that they use is neither expected to be monotone nor submodular.

Based on our new submodular scoring function, we present our  model-agnostic attribution algorithm~\textbf{Submodular Ensembled Attribution for Neural Networks (SEA-NN)}. This algorithm scores features according to their marginal score gains in presence of other critical features, rather than attributing according to the raw scores. For a given input-output pair on a given model, our proposed methodology assigns attribution scores based on the submodular scoring function learned using heatmaps of multiple algorithms. This encourages specificity as the redundant, yet discriminatory, features get lower attribution scores. 
One could also view our methodology as combining multiple attribution algorithms for a given DNN model which is especially useful as the goodness of attribution scores in existing methods can vary based on one's perspective.
Each such attribution method has different biases in defining what may or may not be important in an input. Moreover, \cite{rieger} showed that aggregating attribution maps also results in a more robust attribution map, as adversarial attacks do not transfer well across different attribution methods. Additionally, if each of our input attribution method satisfies the axioms of attribution defined in \cite{pmlr-v70-sundararajan17a}, our method either inherits them or can be easily modified to satisfy most of them. Importantly, we note that the computational overhead of learning the submodular score and the ensembled attribution is not prohibitive and is limited to a small number of evaluation and backpropagation calls, which can be efficiently implemented. More discussion of our method's computational effort is discussed in Section \ref{proposed}, with quantitative analysis in the supplementary materials.


The key contributions of our work can be summarized as follows: 
\vspace{-7pt}
\begin{itemize}
\setlength\itemsep{-0.1em}
\item To achieve the aforementioned objectives of ensembling known attribution heatmaps, we propose a new formulation for learning a submodular set function from real-valued heatmaps. Typical supervision studied in literature for learning submodular functions is of the form of a collection of ``important'' subsets rather than real-valued importance maps. While proposed in the context of ensembling attribution maps, the proposed formulation may be of independent interest in other applications of submodular learning.
\item We propose a new Submodular Ensembled Attribution for Neural Networks (SEA-NN) algorithm, which improves over existing attribution methods in several ways -- importantly, their specificity. Furthermore, this method provides a means to consider the marginal gain of a pixel's attribution in the presence of another pixel (or set of pixels), an aspect that is generally ignored in existing attribution methods.
\item We conducted a comprehensive suite of experiments on standard benchmark image datasets such as FMNIST, CUB, Tiny Imagenet as well as the Brain Tumor Detection dataset where specificity is highly desirable. Our experiments and ablation studies validate the usefulness of the proposed method from a qualitative as well as quantitative perspective. 
\end{itemize}





\section{RELATED WORK}
\label{rel}
\vspace{-8pt}
The last few years have seen a significant number of efforts in enhancing the explainability of machine learning methods -- in particular, the predictions of DNN models. These efforts include local and global methods, model-agnostic and model-specific methods, causal and non-causal methods, and so on \cite{molnar2019,xai2021tutorial,tjoa2020xaisurvey}. In this work, our efforts are based on gradient-based attribution methods, which continue to be among the most popular family of methods, especially on image-based problems. A brief overview of such methods is presented below.

Vanilla Gradient (VG) \cite{DBLP:journals/corr/SimonyanVZ13} and Deconvnet \cite{DBLP:journals/corr/ZeilerF13} were among the earliest methods proposed for attribution in images. Many methods developed upon the notion of gradient-based attribution including Guided Backprop \cite{gbp}, Grad-CAM \cite{selvarajugrad} and Grad-CAM++ \cite{gradcamp}, which introduced newer ways of computing or using the gradient to get class-discriminative attribution maps.
Integrated Gradient (IG) \cite{pmlr-v70-sundararajan17a} was a seminal effort that improved upon gradient-based methods by varying the input $x$ on a linear path from baseline $\bar{x}$ to $x$ and averaging gradients obtained at all these inputs. In order to produce less noisy attribution maps, Smooth Integrated Grad (SG) \cite{smilkov2017smoothgrad} averaged attribution maps of IG corresponding to inputs perturbed with Gaussian noise. Several other attribution methods were introduced. Layer-wise Relevant Propagation (LRP) \cite{Bach2015OnPE} and DeepLIFT \cite{shrikumar2017learning} proposed attributions with discrete gradients using a modified form of backpropagation. There has also been some interesting study on relationships between these methods. \cite{ancona2018better} showed that LRP-$\epsilon$ rule is related to $input\circ gradient$ (element-wise product of input and gradient) and IG is also strongly related to methods like $input\circ gradient$ and DeepLIFT.

However, evidently, each of the aforementioned methods focuses on its own perspective to gradient-based attribution, with its own biases. Very little work has been carried out in ensembling such attribution methods. \cite{rieger} proposed a simple pixel-wise averaging of attribution maps; \cite{NIPS2017_6993} used a notion of weak submodularity to select top-$k$ important pixels in a streaming setting; \cite{bhatt2020evaluating}  aggregated attribution maps from the viewpoint of Shapley values, and focused on instance-based aggregation from a data point's neighbors. The objectives of each of these efforts were different from ours -- \cite{rieger} is the closest to ours, which we compare with in our experiments.

In this work, we learn a new submodular scoring function using heatmaps of given known gradient-based attribution methods. The use of submodularity -- and thus marginal gains -- in our attribution algorithm provides context-dependent attribution scores (i.e. score of a pixel attribution given other pixels) which differentiates it from most existing methods that compute attribution scores independently for each feature. State-of-the-art DNNs are designed to leverage feature inter-dependencies which are not considered in most popular gradient-based methods. \cite{archipelago} recently developed a framework to show how interactions between features attribute to the predictions of DNNs, supporting the need for such a context-dependent attribution score.
In terms of enhancing specificity of attribution maps, \cite{fine-grained} proposed a method for fine-grained visual explanations using an adversarial defense technique. Our objectives are once again different, and our formulation of learning the submodular scoring function can be of independent interest; besides, such a method can also provide an input heatmap for our proposed SEA-NN method.

\section{PROPOSED METHODOLOGY}
\label{proposed}
\vspace{-8pt}
\paragraph{Background.}
Our work builds upon properties of submodular functions. Submodular functions are a special kind of discrete functions, characterized by the diminishing returns property, and appear naturally in many discrete maximization problems like clustering, sensor placement and document summarization \cite{krause2014submodular}. 
For a set function $f:2^V \rightarrow \mathbb{R}$ defined on a ground set $V$, the marginal gain on adding an element $e$ to the set $A$ can be defined as $f(e| A)= f(A\cup e)-f(A)$. $f$ is said to be \emph{submodular} if for any $e \notin B$, for all $A \subseteq V$ and $B \subseteq V$ such that $A \subseteq B$, $f(e| A) \geq f(e| B)$ i.e. the smaller set has a larger gain upon addition of a new element. On the other hand, if both sets have equal marginal gain i.e. $f(e| A) = f(e| B)$, then $f$ is said to be \emph{modular}. In most applications where $f$ acts as a valuation function, $f$ is desired to be non-negative ie. $f(A)\geq 0$ for all $A\subseteq V$. Additionally, if $f(A) \leq f(B)$ for all $A \subseteq B$, then $f$ is said to be \emph{monotonic}.

Given a neural network with $n$ input features, we assume that we have access to $m$ attribution algorithms that are faithful to it. We denote the heatmap (with normalized values in $[0,1]$) obtained by the $i^{th}$ algorithm on a given input/image as $\calH_i\in[0,1]^n$. We assume that the algorithms are reasonable baselines (of which we have quite a few methods today such as Integrated Gradients or SmoothGrad); hence, in general, it can be assumed that the features with high attribution values are indeed critical for discriminatory power of the network on a given input image.

\vspace{-8pt} \paragraph{Learning the Scoring Function.}
In order to ensemble the provided attribution maps, we first define a scoring function, $f: 2^V \rightarrow \mathbb{R}^+$, where $V$ is the set of pixels. This function assigns a non-negative score to a subset of pixels chosen from the input image consisting of $n$ pixels. We denote $f$'s real-valued extension by $f_{ext}:[0,1]^n\mapsto\mathbb{R}^+$ that scores attribution heatmaps. The desirable properties of such a scoring function, given $m$ attribution heatmaps as input, are: (1) The score for pixels belonging to baseline heatmaps must be high; (2) The score for a heatmap with high specificity/selectivity is high, i.e. further increasing attribution values in a selective heatmap does not increase the score significantly; and (3) A discriminatory feature's value in presence of other useful features must be lower than that in their absence. Note that properties (2) and (3) are critical, as attributions based on such scoring functions will promote heatmaps with high specificity and low redundancy. Here, we propose to learn an appropriate scoring function for this purpose.

While ensuring property (1) above is straight-forward, properties (2) and (3) suggest that the proposed scoring function should satisfy the diminishing returns property. Accordingly, we model $f$ by monotone submodular functions and $f_{ext}$ by monotone concave functions. 
This is because submodular functions and their concave extensions are known to accurately model the principle of diminishing returns \cite{iyer2020isit}. And together with monotonicity, such functions allow us to satisfy Property (2) and facilitate early enough saturation of values. Another pragmatic reason for making this choice is that models inducing monotone submodular functions have been well-studied in recent years~\cite{NIPS2016_6361}. This work~\cite{NIPS2016_6361} proposed a class of functions known as Deep Submodular Functions (DSFs), which can be efficiently learned using appropriate neural network training techniques. We leverage this approach to obtain a submodular score function for our work. (We provide more details on DSFs in the supplementary section.)

Let $f_\bw$ denote a DSF learned as a neural network (NN), parameterized by network parameters, $\bw>0$. As shown in~\cite{NIPS2016_6361}, $f_\bw$ restricted to binary-valued inputs, $\{0,1\}^n$, is indeed a monotone submodular function. We also use non-negative, increasing, concave activation functions in our neural network, similar to ~\cite{NIPS2016_6361}. More details of the architecture details of this NN are provided in Sec \ref{expts} and the supplementary section. We now make a critical observation that when such an $f_\bw$ is extended to the domain $[0,1]^n$, it is indeed a concave extension of this submodular function. This is because the activation functions in the network defining the DSF are all non-negative, monotone (increasing), and concave functions. We hence propose to learn the parameters $\bw$ by solving the following novel (intermediate) formulation:
\begin{equation}\label{eqn:int1new}
\begin{split}
    \min_{\bw\ge0} \ \ \frac{\lambda}{2}\|\bw\|^2 + \sum_{i=1}^m\left[f_\bw(\calH^*)-f_\bw(\calH_i)\right] + (1- f_\bw(\calH^*))^+
\end{split}
\end{equation}
where $(a)^+ \equiv \max (0, a)$, $\lambda>0$ is a regularization hyperparameter, and $\calH^*$ is a heatmap with all attributions as unity. This NN-based DSF is obtained by training using the $m$ attribution maps as input. Since $\calH^*$ is the heatmap where any monotone function is maximized, the second term in the objective of Eqn (\ref{eqn:int1new}) promotes functions/parameters that give high score to pixels in all baseline heatmaps. As a result, the scoring function, $f_\bw$ will saturate at heatmaps that resemble the most specific heatmaps among the given attribution maps that we ensemble. The last term in the objective of Eqn (\ref{eqn:int1new}) prevents the trivial solution of $\textbf{w}=\textbf{0}$.

Since $m$, the number of input attribution maps, can be limited in practice , we include further supervision in order to improve the generalization of Eqn (\ref{eqn:int1new}). To this end, we assume that each input heatmap is binarized using $k$ different hard thresholds (resulting in $k \times m$ inputs to learn the DSF). We denote such binarized heatmaps as $\calH_i^j, i\in\{1,\ldots,m\}, j\in\{1,\ldots,k\}$. Let $B_i^j$ denote the number of features selected in $\calH_i^j$ by the binary thresholding. We now propose our final formulation for learning the scoring function as:
\begin{equation}
 \begin{split}
\label{eqn:proposed}
    \min_{\bw\ge0} \ \ \frac{\lambda}{2}\|\bw\|^2 + \lambda_1\Bigg(\sum_{i=1}^m \bigg(f_\bw(\calH^*)-f_\bw(\calH_i)\bigg)\Bigg)+ \\
    \lambda_2\Bigg(\sum_{i=1}^m \sum_{j=1}^k \bigg(\delta + \max_{|A|\le B_i^j} f_\textbf{w}(A)-f_\textbf{w}(\calH_i^j) \bigg)^+ \Bigg)
 \end{split}
\end{equation}
where $\left(a\right)^+\equiv\max(0,a)$; $\lambda>0$, $\lambda_1>0$ \& $\lambda_2>0$ are hyperparameters; and $\delta>0$ also a hyperparameter controlling the margin. In practice, it is possible to have $(\delta+\max_{A : |A|\leq B_i^j}f_\textbf{w}(A)-f_\textbf{w}(\mathcal{H^*}))^+<\delta$ when we are solving the maximization using the approximate greedy algorithm. Hence, we did not need the term to prevent $\textbf{w}=\textbf{0}$ as a trivial solution in our implementation. The new loss terms in Eqn (\ref{eqn:proposed}) promote scores such that the binarized heatmaps are scored high among maps with the same number of selected features. Note that the term $\max_{|A|\le B_i^j} f_\textbf{w}(A)-f_\textbf{w}(\calH_i^j)$ may turn out to be negative whenever the submodular maximization is solved approximately (e.g., when solved via the greedy algorithm). Hence the truncation based on the $(\cdot)^+$ is necessary in the above formulation (Eqn \ref{eqn:proposed}).

To the best of our knowledge, formulations like Eqn (\ref{eqn:proposed}) that jointly learn the submodular function as well as its concave extension are not well-studied in literature. We hence hope that the proposed formulation may be of independent interest in other applications (e.g. weighting samples instead of selecting samples in active learning, imbalanced classification or long-tailed recognition), where early saturating functions need to be induced for similar problems beyond attribution maps. 

The proposed formulation (\ref{eqn:proposed}) can be solved using Mirror Descent or Projected Subgradient Descent. Any subgradient of the objective in Eqn (\ref{eqn:proposed}) is given by: $\lambda\bw+\sum_{i=1}^m[\nabla_\bw f_w(\calH^*)-\nabla_\bw f_w(\calH_i)+\sum_{j=1}^k\rho_{i}^j(\nabla_\bw f_\textbf{w}(\bar{A}_i^j)-\nabla_\bw f_\textbf{w}(\calH_i^j))]$, where $\bar{A}_i^j$ is a solution to the inner submodular maximization problem subject to a cardinality constraint, and $\rho_i^j>0$ iff $\delta+\nabla_\bw f_\textbf{w}(\bar{A}_i^j)-\nabla_\bw f_\textbf{w}(\calH_i^j)>0$ and $\rho_i^j=0$ otherwise. We solve for $\bar{A}_i^j$ using the well-known constant-factor greedy approximation algorithm for maximizing a non-negative monotone submodular function~\cite{DBLP:journals/mp/NemhauserWF78}. This subgradient can be efficiently computed by backpropagating through the NN modeling the DSF.

Notably, the function learnt, $f_\bw$, exhibits certain degree of stability to noise. This is because $f_\bw$, when restricted to the domain $\{0,1\}^n$, is a non-negative, monotone, submodular function. Such functions are known to be noise-stable~\cite{noise-stable}. Also, on the compact domain $[0,1]^n$, $f_{ext}$ is a concave function and hence is Lipschitz continuous (for e.g., see theorem~C.4.1~in~\cite{Ne05}). Hence the proposed scoring function is expected to be fairly robust to noise in baseline attribution maps.

\begin{lemma} Given input $x$ and a classifier neural network function $F^c$ corresponding to class $c$, $|f_{ext}(\calH_i(x, F^c))-f_{ext}(\calH_i(x+\delta, F^c))| \leq L||\calH_i(x, F^c)-\calH_i(x+\delta, F^c)|| ~\forall \calH_i, i\in\{1,\ldots,m\}$, $L$ being a constant.
\end{lemma}

This lemma highlights that the deviation in $f_{ext}$ caused by an additively perturbed input is not arbitrarily high but depends on the deviations that arise from the baseline heatmaps.

We now discuss our algorithm to compute a new heatmap that ensembles baseline heatmaps using this induced scoring function.

\vspace{-8pt} \paragraph{Submodular Ensemble Attribution Algorithm.}
We begin by noting that high specificity in attribution heatmaps may not be easy to achieve by arbitrary linear/non-linear combinations of the given heatmaps (due to the interactions between pixels, as stated in Sec \ref{rel}). 
Thus we propose to learn $f_{\bw}$ that attributes features based on their marginal gain in the context of other features (i.e. $f_{\bw}$ scores a subset of pixels rather than each individual pixel, which is a key difference from a linear combination of these heatmaps). The proposed attribution algorithm is detailed in~Algorithm~(\ref{alg:Attribution-Algorithm}). Our algorithm is designed with the following key features/advantages: (i) It improves as the given heatmaps that we ensemble improve; (ii) It is expected to achieve high specificity; and (iii) By inheritance, the axiom of ``Implementation Invariance'' defined in~\cite{pmlr-v70-sundararajan17a} is satisfied whenever the given heatmaps satisfy the same axiom.  

Many types of other prior information can also be easily encoded as constraints in the proposed formulation (Eqn \ref{eqn:proposed}). In particular, one can insist that the proposed attribution satisfies the axioms Sensitivity(a) and/or Sensitivity(b)~\cite{pmlr-v70-sundararajan17a} by simply adding appropriate constraints in Eqn (\ref{eqn:proposed}). For e.g., if the $i^{\text{th}}$ feature needs to have non-zero attribution, one may include the constraint: $f_\bw(\mathbf{1})-f_\bw(\mathbf{i})\ge\epsilon$, where $\mathbf{1}$ is vector of all ones, and $\mathbf{i}$ is vector of all ones except that the $i^{\text{th}}$ entry is zero, and $\epsilon$ is a small tolerance. If this constraint is satisfied, the marginal gain of this feature w.r.t. any subset of features and hence the attribution value will be greater than or equal to $\epsilon$, again because of the diminishing returns property. Analogously, if a feature, say $v$, must have zero attribution, then adding the constraint $f_\bw(\{v\})=f_\bw(\mathbf{0})$ should suffice, where $\mathbf{0}$ is the vector of all zeros. Lastly, the proposed methodology offers some unique features like: (a) one can attribute a set of images (say, all examples of a particular class) by simply pooling all the attributions of all the images and learning the submodular score. This will be interesting because now features important at set (class) level will be revealed; (b) If one needs to find $p$ features that are most influential, then instead of picking the top-$p$ attributed features, we can now solve for $\arg\max_{A\subseteq\{1,\ldots,n\},|A|\le p}f_\bw(A)$. This would give the top $p$ discriminatory, yet non-redundant set of features.

\begin{algorithm}[t]
\footnotesize
   \caption{Attribution Algorithm}
   \label{alg:Attribution-Algorithm}
\begin{algorithmic}
   \STATE {\bfseries Input:} Trained DSF $f$, Set of features to attribute $V$
   \STATE Initialize feature subset $A=\{\}$
   \STATE Initialize $n=|V|$
   \STATE Initialize attribution map $G[i]=0$ for $i=0,1,\ldots,n$
   \STATE $M=\{\arg\max_{v\in V\setminus A} f(v|A)\}$, Pick $e\in M$    
   \WHILE{$|A|<n$ and $f(e|A)>0$}
   \FOR{$p \in M$}
   \STATE $G[p]=f(e|A)$
   \ENDFOR

   \STATE $A=A\cup \{e\}$; $V=V\setminus M$
   \STATE $M=\{\arg\max_{v\in V}
   f(v|A)\}$, Pick $e\in M$
   \ENDWHILE
   \RETURN $G$
\end{algorithmic}
\end{algorithm}

\begin{figure*}[h!]
    \centering
    \includegraphics[width = \textwidth]{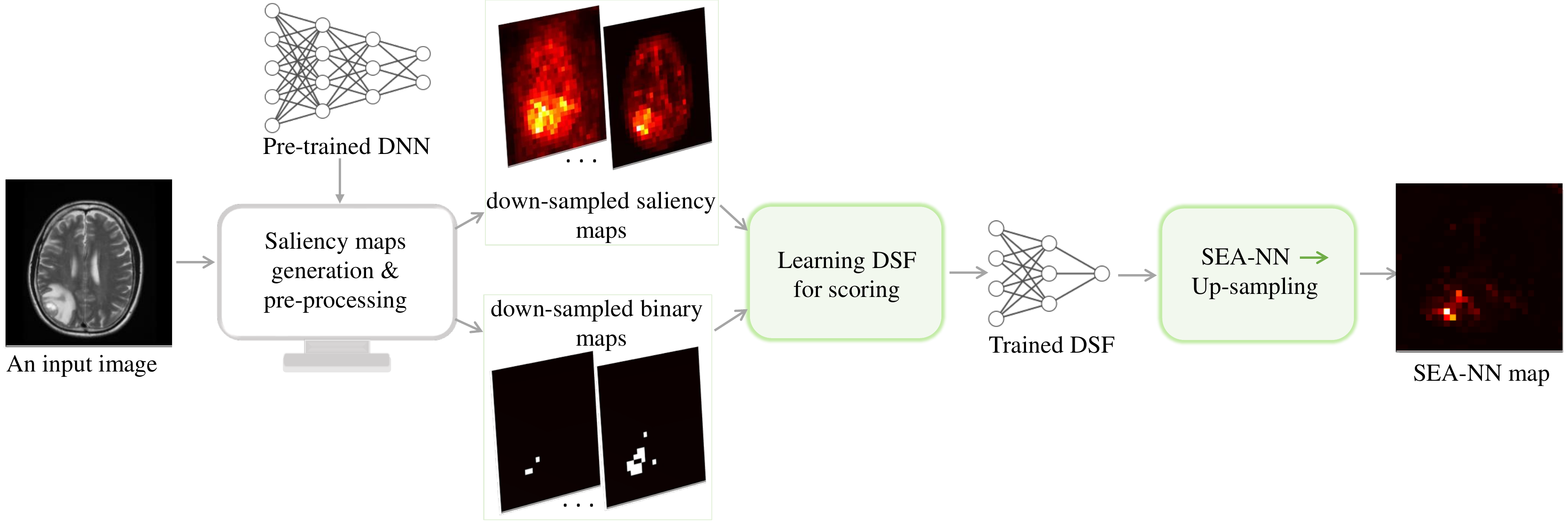}
    \vspace{-12pt}
    \caption{Overview of SEA-NN for the more general high-dimensional setting}
    \vspace{-8pt}
    \label{fig:workflow}
\end{figure*}

\vspace{-11pt} \paragraph{Computational Effort.} The main computational overhead for the projected subgradient algorithm is the computation of $\bar{A}_i^j\ \forall i\in\{1,\ldots,m\}\ \forall\ j\in\{1,\ldots,k\}$. Interestingly, the use of the greedy algorithm ensures that computing $\bar{A}$ for larger thresholds gives those for the smaller ones too. Hence at every iteration of the descent, only one call to the greedy algorithm is required with cardinality bound as $B\equiv \max_{i\in\{1,\ldots,m\}, j\in\{1,\ldots,k\}}B_{i}^j$, leading to $O(Bn)$ evaluations of DSF output. Typically, $B\ll n$, thus leading to a marginal overhead of $O(n)$ DSF evaluations.
In implementation, our module for cardinality-constrained submodular maximization performs $O(B)$ calls to the DSF where in each call, we pass $n$ inputs as a single batch. (We report quantitative measures of time complexity in the supplementary section.)

\vspace{-8pt} \paragraph{Scaling to High-Dimensional Images.}\label{high-d}
For high-dimensional images, we learn a DSF using down-sampled versions of heatmaps and their binary (thresholded) counterparts. The proposed SEA-NN algorithm then computes attribution scores using the learned DSF. The attribution map thus obtained is upsampled by interpolation (we use simple nearest-neighbor interpolation in this work) to match the original resolution of the image. Similar kind of upsampling techniques are also part of widely used attribution algorithms like CAM \cite{zhou2015learning} and Grad-CAM \cite{selvarajugrad}. For the downsampling itself, we divide a heatmap into grids and represent each grid by a single pixel in the downsampled heatmap where we assign average of attribution scores in the grid as the corresponding pixel's value in the downsampled version. For downsampling the binary maps, we divide the binary map into grids, and each grid cell's value is set to one if the sum of values inside it is greater than the average of the sums of values across all grid cells in the images. This provides a sense of relative importance for each grid location. An illustration is provided in the supplementary section for clarity. Empirically, however, we found that our method was not sensitive to the specifics of upsampling and downsampling.

An overall workflow of our approach for the more general high-dimensional setting is shown in Fig \ref{fig:workflow}.

\section{EXPERIMENTS AND RESULTS}
\label{expts}
\vspace{-8pt}
In order to validate the proposed method, we performed a comprehensive suite of experiments on the Fashion-MNIST (or FMNIST) \cite{xiao2017fashionmnist}, CUB \cite{cub}, Tiny Imagenet \cite{Le2015TinyIV} and Brain Tumor Detection \footnote[1]{https://www.kaggle.com/navoneel/brain-mri-images-for-brain-tumor-detection} datasets. Following \cite{pmlr-v139-lu21b}, we used Vanilla Gradient (VG), Integrated Gradients (IG) and Smooth Integrated Gradient (SG) as the baseline attribution maps for our experiments in this section. These methods are widely used across DNN architectures, and were also found to be sufficient to learn a submodular scoring function that gave good results. We show more results with other methods in the supplementary section.

\noindent \textit{Baselines and Evaluation Metrics:} We show qualitative and quantitative comparison against VG, IG, SG, as well as Agg-Mean \cite{rieger} that does linear averaging of attribution maps. (Due to unavailability of source code, we are unable to compare against \cite{fine-grained}, which however did not have the objective of ensembling attribution maps, but provided fine-grained explanations.) In addition to showing the qualitative results of the heatmaps, we also studied the methods quantitatively using the Area Under Perturbation Curve (AUPC) metric \cite{7552539}. We perturb regions in the input image cumulatively, in the decreasing order of their relevance where relevance of a region depends on the attribution score assigned to it by an attribution method. The classifier's score corresponding to the predicted label is recorded after each perturbation forming a perturbation curve. If an attribution method correctly highlights the relevant regions in an image, there will be a sharp decrease in the classifier's score, making the area under the perturbation curve low. We report the AUPC metric averaged across all inputs. We follow the implementation of AUPC as suggested in \cite{GohLWSB21Understanding} and \cite{Petsiuk2018rise}.

\noindent \textit{Implementation Details:} 
We used PyTorch \cite{pytorch} for all our implementations, and used the Captum library \cite{captum2020github} to obtain the heatmaps for learning our DSF. In order to obtain an attribution map for a specific input (local explanation), we learn a DSF per input. The DSF architecture was chosen as Linear(784, 512)$\rightarrow$Linear(512, 256)$\rightarrow$Linear(256, 32)$\rightarrow$Linear(32, 1) with square-root as the activation function after all but the last layer. The dataset for training a DSF comprised of heatmaps from VG, IG, SG along with their binary counterparts obtained using 10 threshold values equally spaced between 5 and 50. As our DSF is a small neural network with very few parameters, this size of training data was empirically found sufficient. Weights of the DSF NN were initialized to 1 and training was done for 50 epochs. For training, we used Adagrad optimizer with learning rate decay coefficient as 0.1. The hyperparameter $\delta$ was set to $10^{-5}$.  For the high-dimensional setting (CUB, TinyImagenet and BTD datasets), the downsampled version of SEA-NN map was chosen to be of resolution $28 \times 28$. Other hyperparameters specific to a dataset are mentioned while describing the corresponding experiments.

\begin{table}
\caption{Area Under Perturbation Curve averaged over images of the datasets (lower is better)} \label{aupc}
\begin{center}
\begin{tabular}{lllll}
\textbf{Method} &\textbf{FMNIST} &\textbf{BTD} &\textbf{CUB} &\textbf{T.Img}\\
\hline
SEA-NN       & \textbf{3.46} & \textbf{11.11} & \textbf{4.82} &  \textbf{5.42}\\
IG         & 3.62    & 16.78 & 4.9 &  5.49\\
VG         & 3.68    & 17.31 & 5.2 &  5.51\\
SG          & 3.57   & 17.51 & 5.21 &  5.79\\
Agg-Mean    & 3.51  & 17.24 & 4.84 & 5.47  \\
\hline
\end{tabular}
\end{center}
\end{table}

\begin{figure}[ht]
    \centering
    \includegraphics[width = 0.95\columnwidth]{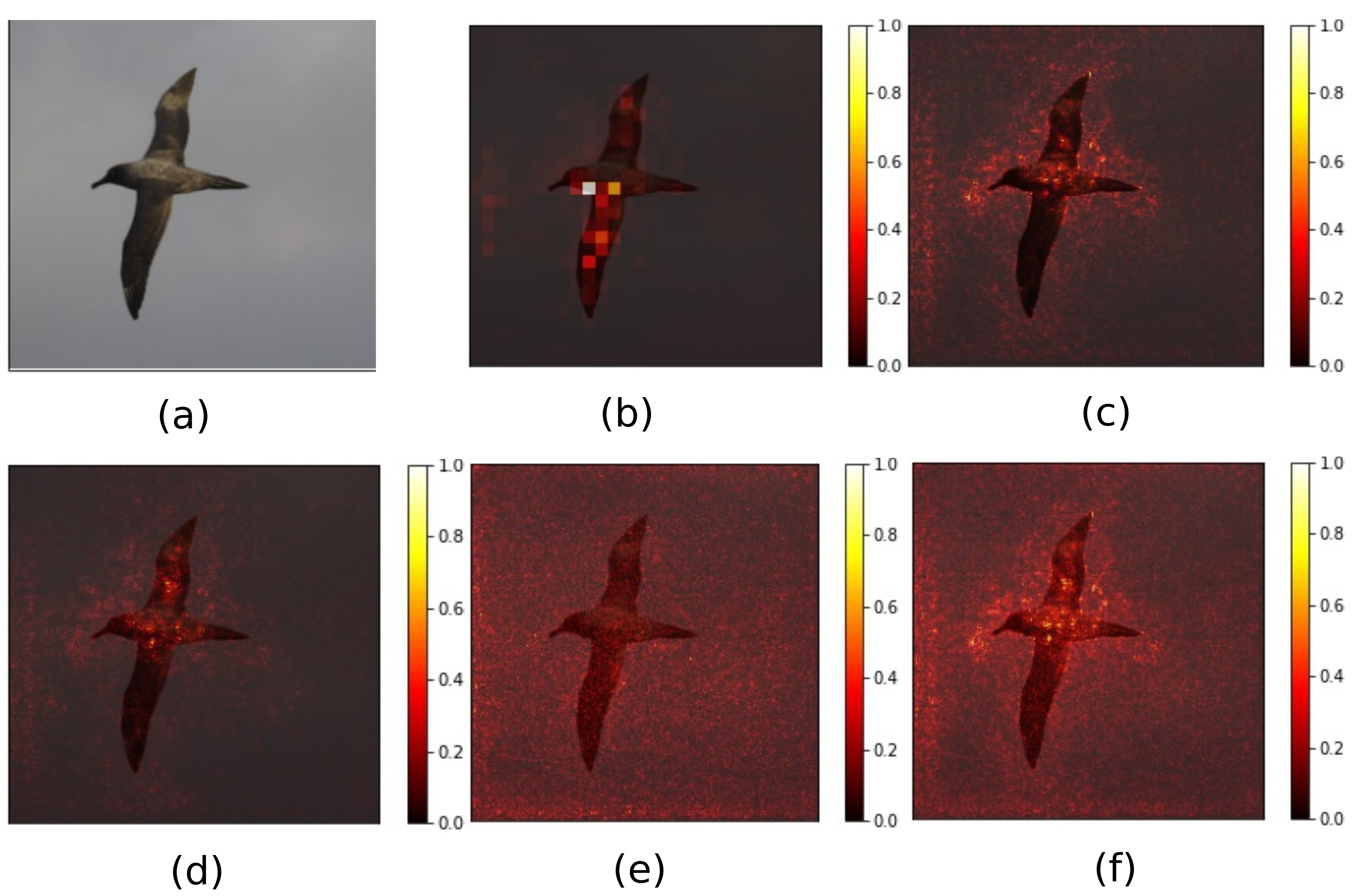}
    \vspace{-10pt}
    \caption{Example image and attribution maps on CUB dataset image: (a) Image of Sooty Albatross (b) SEA-NN (c) Integrated Gradient (d) Vanilla Gradient (e) Smooth Integrated Gradient (f) Agg-Mean}
    \vspace{-6pt}
    \label{fig:cub}
\end{figure}
\vspace{-10pt} \paragraph{Results on CUB.}
We used a fine-tuned Resnet-18 classifier and experiment on the train split of CUB. The hyperparameters were set to: $\lambda_1$ as 0.1, $\lambda_2$ as 10 and Adagrad's weight decay coefficient as $10^{-6}$. From Table \ref{aupc}, we observe that SEA-NN outperforms baseline methods on the average AUPC metric. Fig \ref{fig:cub} shows qualitative results, demonstrating the higher specificity of SEA-NN over baseline methods. More qualitative results are provided in the supplementary section.

\begin{figure}[ht]
\label{fig:boy}
    \centering
    \includegraphics[width = 0.95\columnwidth]{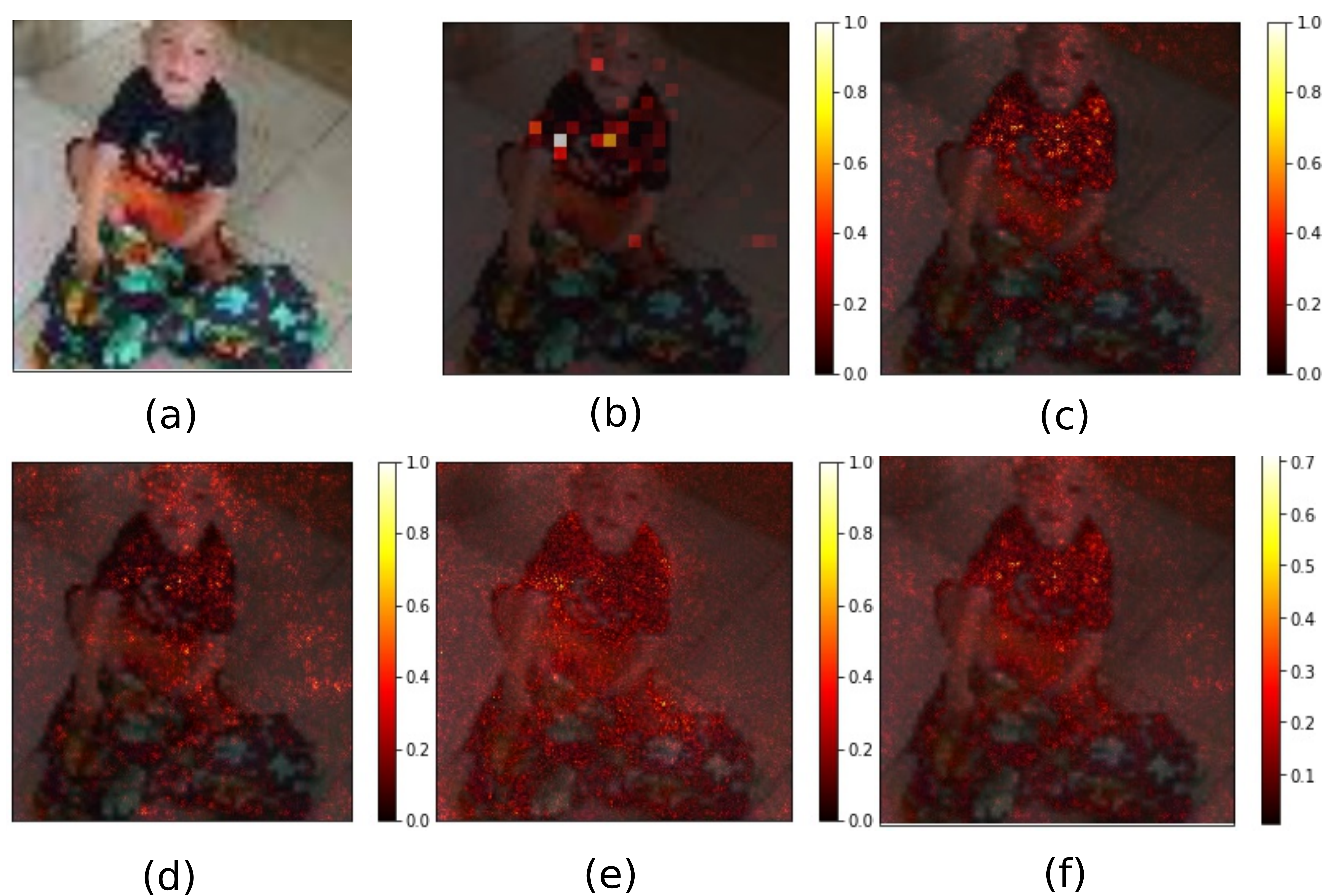}
    \vspace{-10pt}
    \caption{Example image and attribution maps on TinyImagenet dataset image: (a) Image of boy (b) SEA-NN (c) Integrated Gradient (d) Vanilla Gradient (e) Smooth Integrated Gradient (f) Agg-Mean}
    \vspace{-6pt}
\end{figure}

\vspace{-8pt} \paragraph{Results on TinyImagenet (T.Img).} 
A fine-tuned Resnet-18 with top-1 accuracy of 57\% was used as the classifier. The hyperparameters were set to: $\lambda_1$ as 0.1, $\lambda_2$ as 10 and Adagrad's weight decay coefficient as $10^{-6}$. Fig \ref{fig:boy} shows an example where SEA-NN outperforms baseline methods on specificity of the heatmap. As shown in Table \ref{aupc}, the AUPC score averaged across the validation split of Tiny Imagenet for SEA-NN was the best among the considered methods. More qualitative results are in the supplementary section.

\begin{figure}[ht]
    \centering
    \includegraphics[width = \columnwidth]{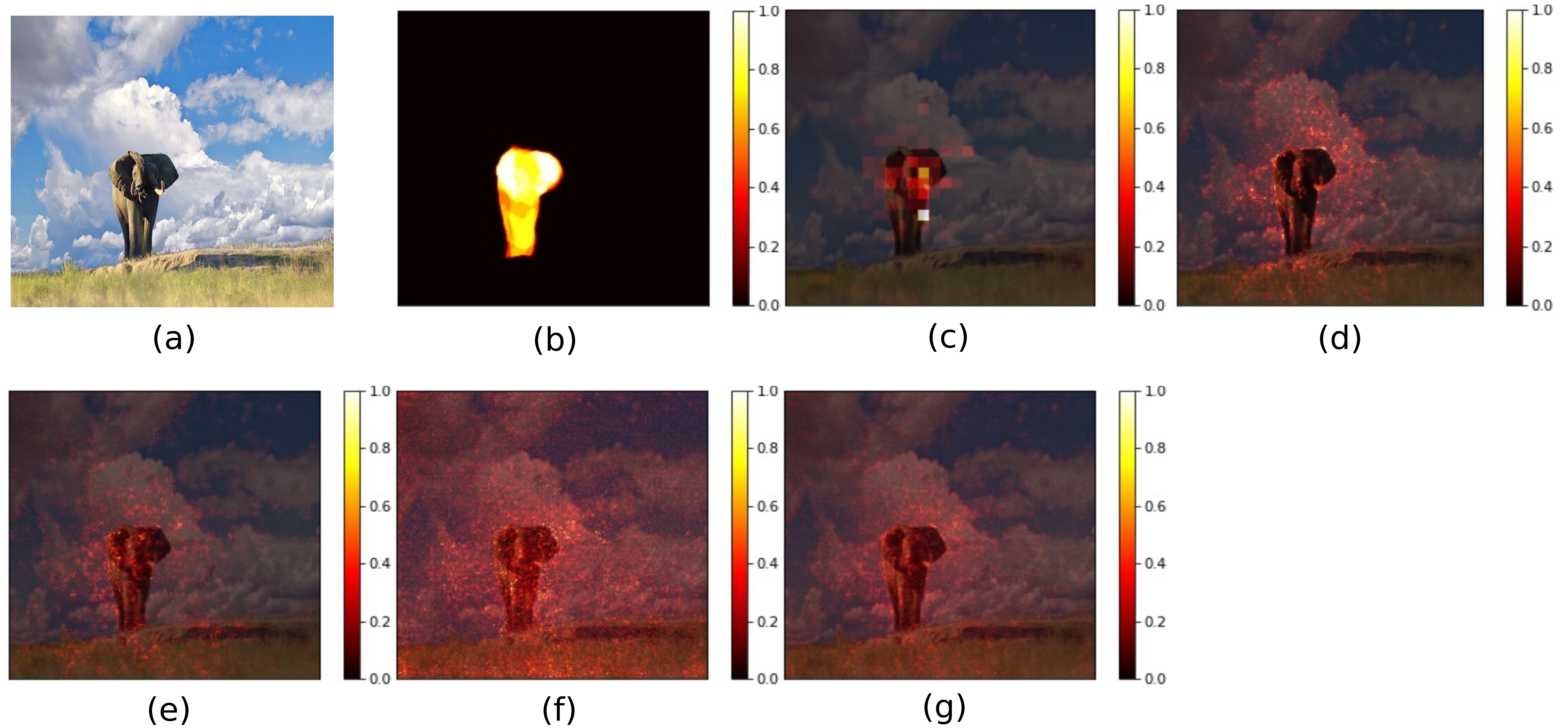}
    \vspace{-15pt}
    \caption{Correlations with human annotated heatmaps experiment (Imagenet image and attribution maps): (a) Image of an elephant (b) Human annotation (c) SEA-NN (d) Integrated Gradient (e) Vanilla Gradient (f) Smooth Integrated Gradient (g) Agg-Mean}
    \label{fig:human-imgn}
\end{figure}

\begin{figure}[ht]
    \centering
    \includegraphics[scale = 0.5]{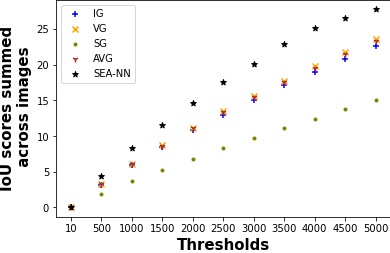}
    \vspace{-10pt}
    \caption{Results on Correlations with human annotations (higher is better)}
    \vspace{-10pt}
    \label{fig:sum-iou}
\end{figure}

\vspace{-8pt} \paragraph{Correlations with human annotations.}
We conducted an experiment to see how well SEA-NN maps align with human annotated heatmaps. For this, we use human annotations provided by \cite{mohseni} for 98 images of the Imagenet dataset. We compute the Jaccard score between an attribution map and the annotated heatmap after hard-thresholding them to keep only the top-$k$ pixels. We compute this for multiple thresholds ($k$ values) and show the results in Fig \ref{fig:sum-iou}. The figure demonstrates that the pixels picked by SEA are the most human-interpretable among all baselines method. We also show this through a qualitative example in Fig \ref{fig:human-imgn}. Hyperparameters were: $\lambda$'s as 0.1 and weight decay coefficient as $10^{-3}$.

\vspace{-8pt} \paragraph{Results on Brain Tumor Detection (BTD).}
We used a fine-tuned VGG-11 classifier with top-1 accuracy of 85\%. In the BTD dataset, white mass inside the skull is known to be indicative of brain tumor. The hyperparameters were set to: $\lambda_1$ as 1, $\lambda_2$ as $100$ and weight decay coefficient for Adagrad as $10^{-3}$. Fig \ref{motiv-ex} shows that SEA-NN correctly identifies the affected regions with minimum noise compared to all baseline methods. Sparsity is highly beneficial in medical diagnosis where false positives could be fatal. Table \ref{aupc} shows that SEA-NN obtains the best average AUPC score.
\begin{figure}\label{fig:trouser}
    \centering
    \includegraphics[width = 0.9\columnwidth]{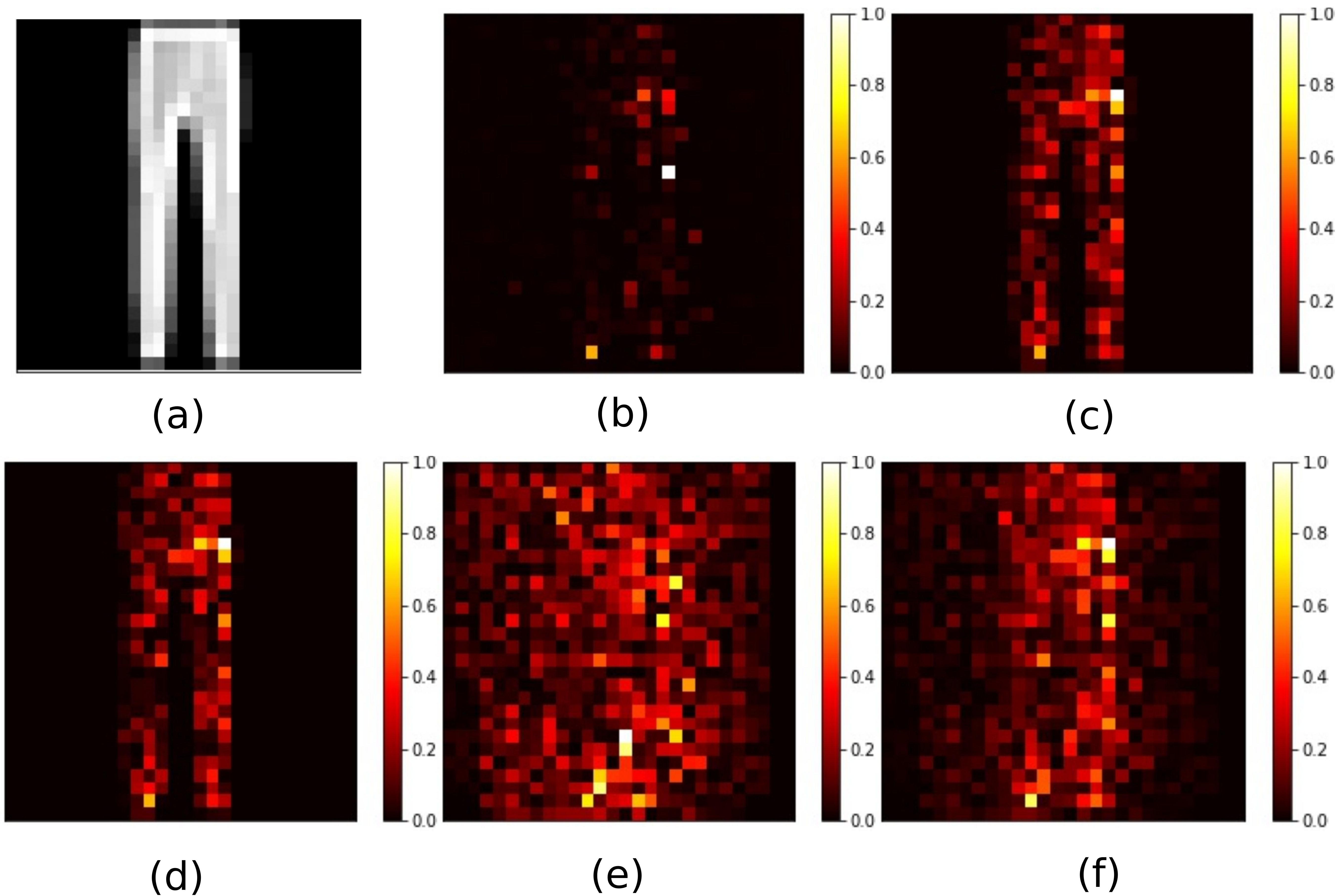}
     \vspace{-12pt}
    \caption{Example image and attribution maps on FMNIST dataset: (a) Image of Trouser (b) SEA-NN (c) Integrated Gradient (d) Vanilla Gradient (e) Smooth Integrated Gradient (f) Agg-Mean}
    \vspace{-6pt}
\end{figure}
\vspace{-8pt} \paragraph{Experiment on Fashion-MNIST (FMNIST).}
We used a simple CNN classifier with accuracy of 90.5\%. The hyperparameters were set to: $\lambda_1$ as 10, $\lambda_2$ as 10 and Adagrad's weight decay coefficient as $10^{-6}$. SEA-NN identified discriminatory pixels to achieve the best AUPC averaged across the test split of FMNIST (Table \ref{aupc}). SEA-NN map shows high specificity (Fig \ref{fig:trouser}). 

\begin{figure}[h!]
\label{fig:top-k}
    \centering
    \includegraphics[width=0.9\columnwidth]{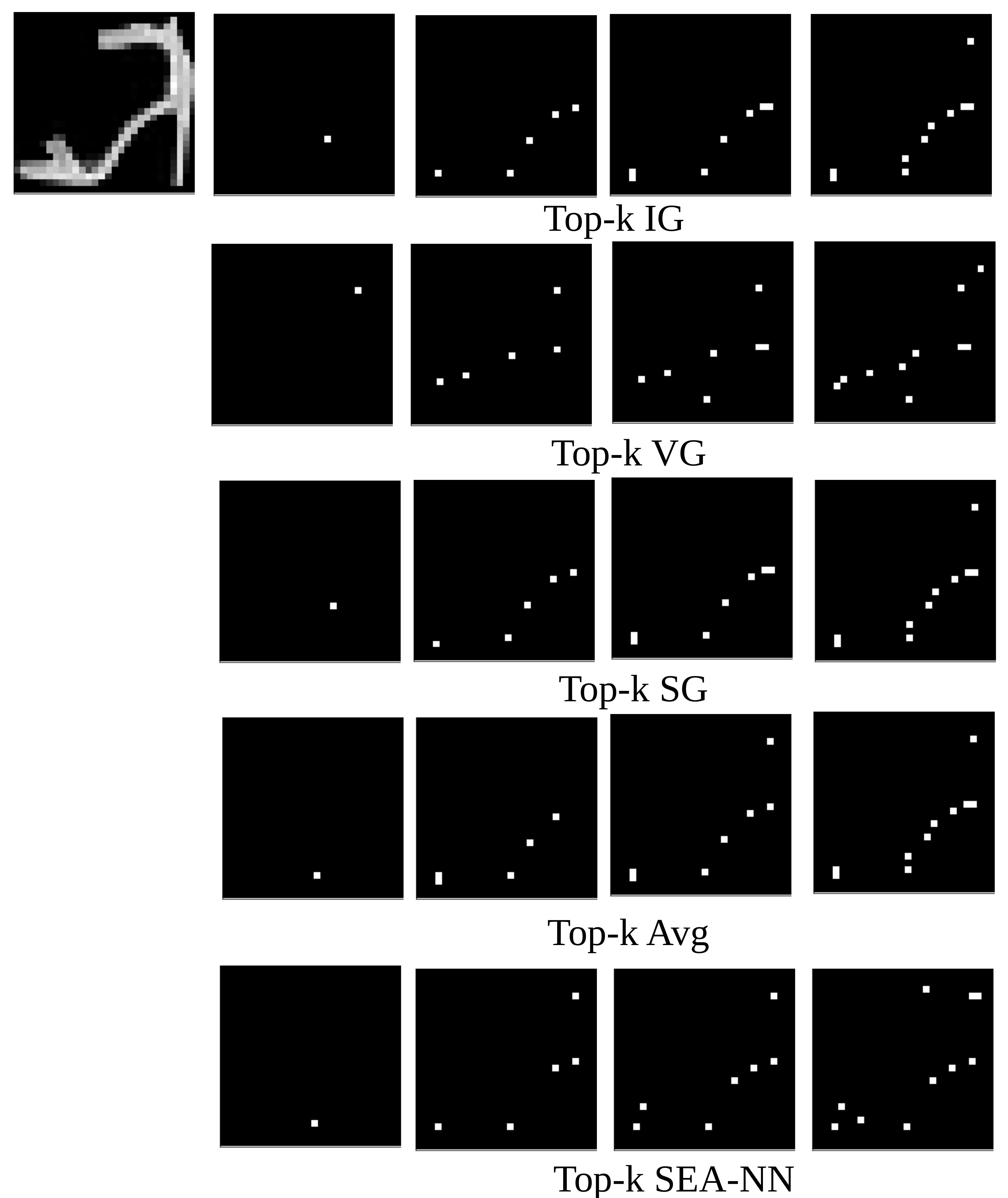}
    \vspace{-10pt}
    \caption{Top-$k$ pixels selected by attribution methods on image from Sandal class in FMNIST. Columns from left to right represent $k$ as 1, 5, 7, 10 respectively}
    \vspace{-6pt}
\end{figure}

\vspace{-8pt} \paragraph{Visualizing Top-K Pixels.}
In order to show the effect of feature-interaction in attribution, we present binary heatmaps showing the top-$k$ pixels identified by various attribution methods on the FMNIST dataset. As shown in Fig \ref{fig:top-k}, the top-$k$ pixels identified by SEA-NN are less redundant. We note that for the same budget, SEA-NN selects a more diverse set of pixels which is especially visible in the rightmost column when $k$ is chosen as 10.

More results -- qualitative and quantitative -- are included in the supplementary section owing to space constraints. 
\vspace{-10pt}
\section{CONCLUSIONS AND FUTURE WORK}
\label{sec_conc}
\vspace{-10pt}
In this work, we propose a framework, SEA-NN, to improve existing attribution methods through submodular ensembling that provides more specificity to the attribution maps, and also context-dependent attribution scores. Ensembling attribution maps of different methods is expected to reduce epistemic uncertainty and biases associated with each attribution method. The special properties satisfied by our learned submodular scoring function and the ease of adding additional constraints during training help making our solution suitable for needs of different users. We validated SEA-NN against well-known gradient-based attribution methods as well as existing ensembling methods and showed that our attribution maps are superior quantatively and qualitatively. We also showed that SEA-NN aligns well with human annotated heatmaps available for a subset of Imagenet images. 

\noindent \textbf{Future Work and Social Impact.}
Attribution methods help enhancing the interpretability of DNNs which will encourage their usage in critical domains such as healthcare and criminal justice. The proposed method can be especially beneficial for domains like medical diagnosis and astronomy where specificity in attribution is highly desired. As a future work, we would like to explore a transfer learning approach such that DSF's trained on just a subset of images, could work well on the entire dataset. This will also help further mitigate the computational effort incurred by our method. 

\subsubsection*{Acknowledgements}
We are grateful to the Ministry of Human Resource Development, India; Department of Science and Technology, India; Honeywell, India for the financial support of this project. We also thank the Japan International Cooperation Agency and IIT-Hyderabad for the provision of GPU servers used for this work. We thank the anonymous reviewers for their valuable feedback. We would also like to thank the AISTATS Mentorship program and our submission mentor, Dr Bilal Alsallakh. PM thanks Google for the PhD Fellowship.

\bibliographystyle{apalike}
\bibliography{ref}


\clearpage
\appendix

\thispagestyle{empty}

\onecolumn \makesupplementtitle

In this supplementary material, we include additional information which could not be included in the main paper due to space constraints. In particular, we include the following:
\begin{itemize}
    \item Description of baseline attribution maps that we use for learning our scoring function
    \item More details on Deep Submodular Functions (DSF)
    \item More details on Lemma 1
    \item Toy illustration for down-sampling binary heatmaps
    \item More results (in continuation to Section 4)
    \item Quantitative comparison of computation time.
\end{itemize}

\section{BASELINE ATTRIBUTION ALGORITHMS}
In this section, we briefly describe the attribution methods using which we learn our submodular scoring function. We use $x$ to denote the input image and $F^c$ to denote neural network's output corresponding to class $c$.
\paragraph{Vanilla Gradient (VG)} VG assigns attribution score for the $i^{th}$ input dimension as

\[\text{VG}_i(x; F^c) := \frac{\partial{F^c}}{\partial{x_i}}\]

\paragraph{Integrated Gradients (IG)} IG tries to mitigate the saturating gradients problem in VG by aggregating the attribution maps corresponding to inputs lying on a path from a baseline $x'$ to the original input $x$.
IG assigns attribution score along $i^{th}$ dimension for input $x$ as

    \[\text{IG}_i(x; F^c):=(x_i-x_i')\int_{\alpha =0}^1 \frac{\partial F^c(x'+\alpha (x-x'))}{\partial x_i} \partial \alpha\]

\paragraph{Smooth Integrated Gradients (SG)} Smooth Grad variants are designed to remove noise from attribution maps by adding noise to the input.
Attribution score for Smooth Integrated Gradients along input dimension $i$ is computed as 

\[\text{SG}_i(x; F^c):= \frac{1}{N}\sum_{j =1}^N 
\text{IG}_i(x+n_j; F^c)\]

where $n_j\sim \mathcal{N}(0, \sigma^2)$ for $j=1 \dots N$. 

\section{MORE DETAILS ON DEEP SUBMODULAR FUNCTIONS}
A Deep Neural Network (DNN) whose weights are restricted to be non-negative and the activation functions used are monotone non-decreasing concave for non-negative reals, constitutes a non-negative monotone non-decreasing submodular function when given Boolean input vectors. This is referred to as Deep Submodular Function (DSF) that can be trained in the similar way as DNNs. The key result that forms the basis of DSF is that sums of concave functions, composed with modular functions (SCMMs), can be easily shown to be submodular. We urge the interested readers to refer \cite{NIPS2016_6361} for more details.

For choosing the NN architecture for DSF in our work, we started with a two-layer NN and kept on adding a layer until our objective over epoch curve started looking saturated at the end of 50 epochs. As described in section 4 of the main paper, the NN architecture chosen for DSF was Linear(784, 512) $\rightarrow$ Linear(512, 256) $\rightarrow$ Linear(256, 32) $\rightarrow$ Linear(32, 1). As the choice of concave activation function, we used $g(x) = \sqrt{x}$ after all but the last layer. We also tried $g(x) = \log(1+x)$ as the activation function but we observed that using square root as the activation function resulted in smoother attribution maps.

\section{MORE DETAILS ON LEMMA 1}
For the DSF architecture used in the paper, the Lipschitz continuity constant ($L$) for $f_{ext}$ (concave extension of DSF) comes out to be $\frac{\Pi_{i=1}^4(\textbf{w}_i^T\textbf{1})^{\frac{1}{2^{5-i}}}}{7}$ where $\textbf{w}_i$ denotes vectorized weights of layer $i$ of the DSF. For deriving this $L$, we applied Theorem~C.4.1~in~\cite{Ne05} and the property that $f_{ext}$ is a monotonically increasing function defined over $[0, 1]^{784}$. 

We also computed $\sup_{\textbf{x}, \textbf{y}} \frac{|f_{ext}(\textbf{x})-f_{ext}(\textbf{y})|}{||\textbf{x}-\textbf{y}||}$ over $10^6$ pairs of random inputs $\textbf{x}, \textbf{y} \sim  [0,1]^{784}$ for a DSF learnt on a random image from the Imagenet dataset and obtained $L$ estimate as 6.54, which is not very high.

\section{DOWN-SAMPLING BINARY HEATMAPS}
\begin{figure}[h!]
    \centering
    \includegraphics[scale=0.3]{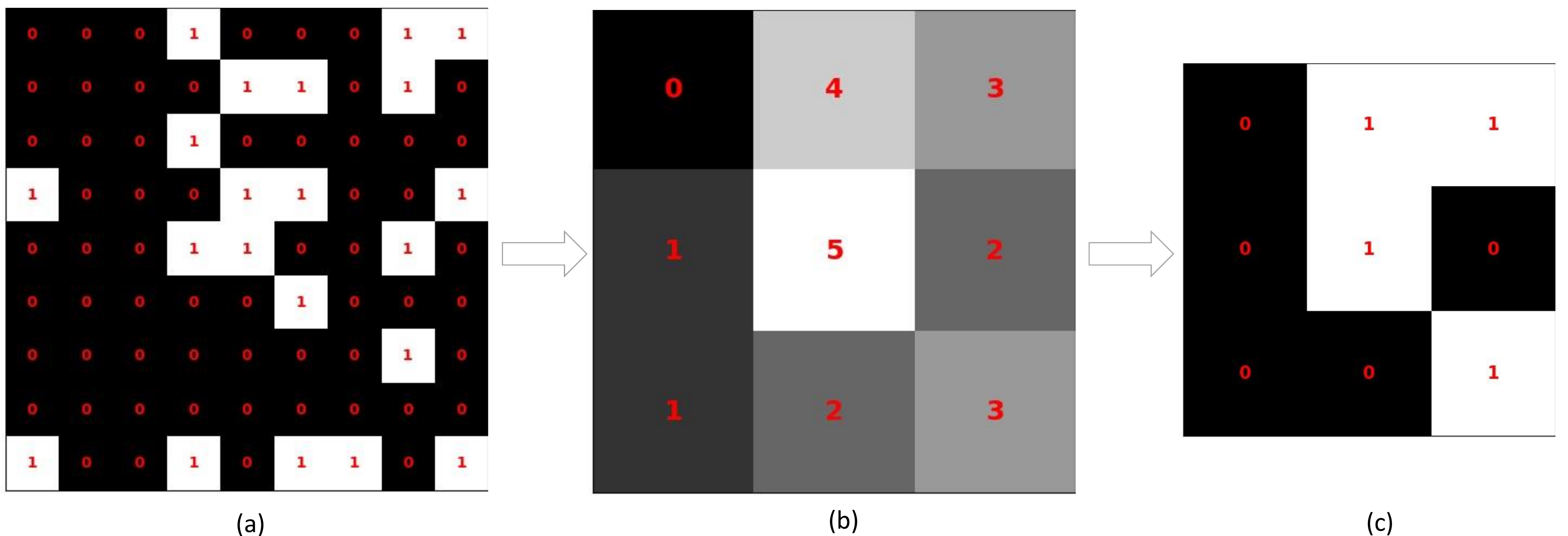}
    \caption{Toy illustration of down-sampling binary heatmaps (a) Binary attribution map (b) Intermediate representation considering 3x3 grids (c) Down-sampled binary map}
    \label{down-binary}
\end{figure}

As detailed in section 3 of the main paper, for the case of high dimensional images, we learn our scoring function using down-sampled versions of baseline heatmaps and their binary (thresholded) counterparts in order to have better computational efficiency. In Fig \ref{down-binary}, we present a toy illustration of the down-sampling procedure for binary heatmaps.

\section{MORE RESULTS}
SEA-NN is designed to enhance specificity in attribution. However, in the absence of standard metrics that simultaneously quantify sensitivity and specificity, we evaluate the discriminative power of our method using the AUPC metric and show qualitative results demonstrating good specificity.

We note that SEA-NN encompasses multiple benefits. It not only improves specificity in attribution but is a novel method to aggregate feature attribution and provides attribution scores that respect feature interactions. Moreover, we propose a new formulation for learning a submodular set function using real-valued inputs. To the best of our knowledge, there is no other method that provides all these benefits. 

Following we demonstrate some more results showing the efficacy of the proposed approach.

\subsection{More Qualitative Results}
\begin{figure}[h!]
    \centering
    \includegraphics[scale= 0.15]{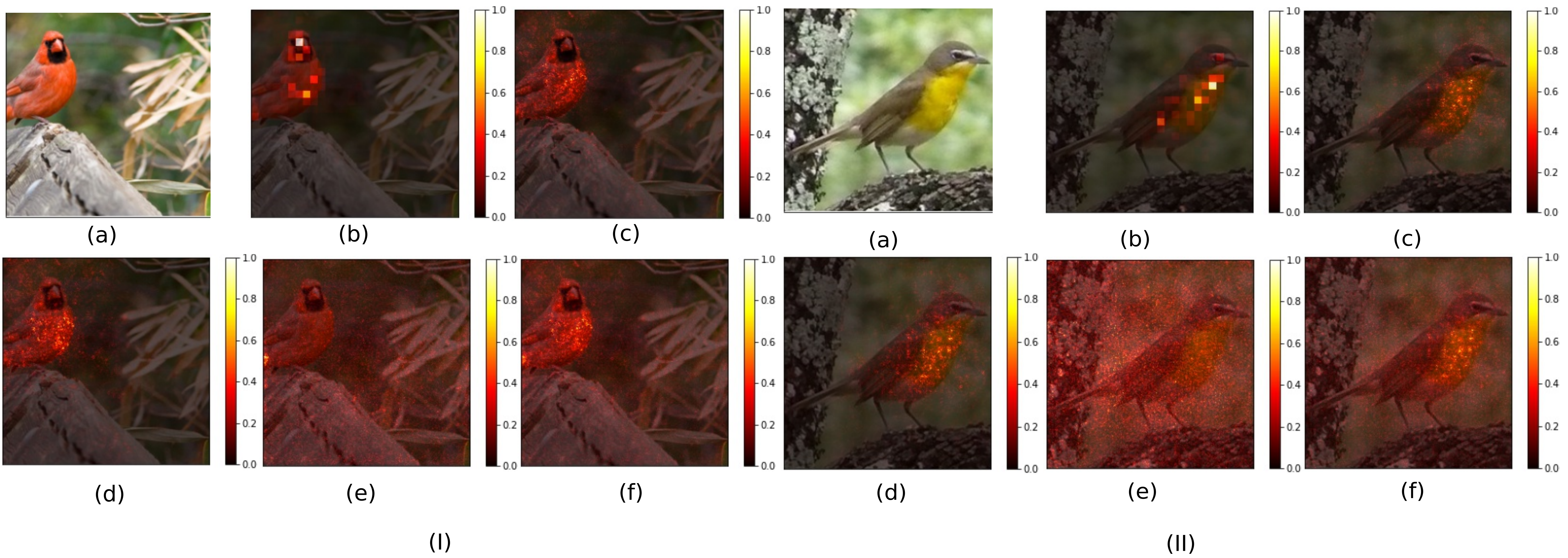}
    \caption{(I) and (II) are images of Golden Retriever birds shown with attribution maps overlaid on the respective images: (a) Image; (b) SEA-NN map; (c) Integrated Gradient map; (d) Vanilla Gradient map; (e) Smooth Integrated Gradient map; (f) Agg-Mean map of (c)-(e) maps.}
    \label{golden-ret}
\end{figure}

\begin{figure}[h!]
    \centering
    \includegraphics[scale=0.15]{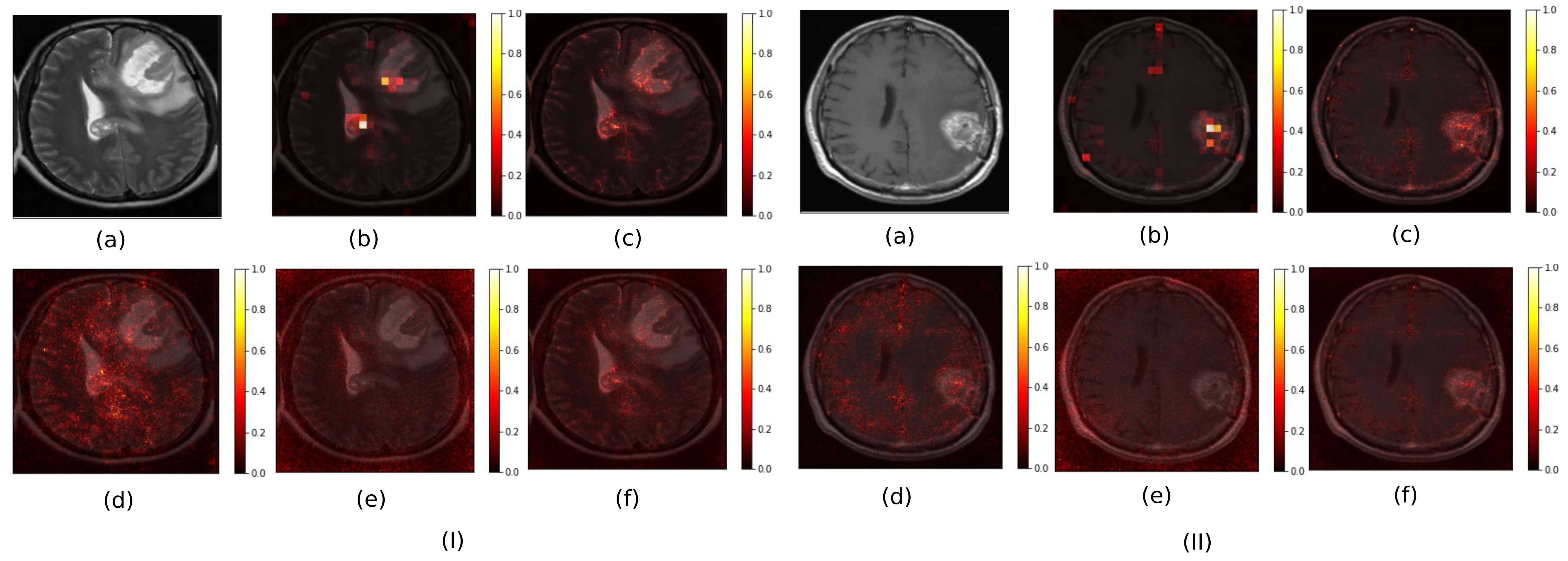}
    \caption{(I) and (II) are images from Brain Tumor dataset shown with attribution maps overlaid on the respective images: (a) Image with brain tumor; (b) SEA-NN map; (c) Integrated Gradient map; (d) Vanilla Gradient map; (e) Smooth Integrated Gradient map; (f) Agg-Mean map of (c)-(e) maps.}
    \label{brain-flat}
\end{figure}

\begin{figure}[h!]
\label{img}
    \centering
    \includegraphics[scale= 0.15]{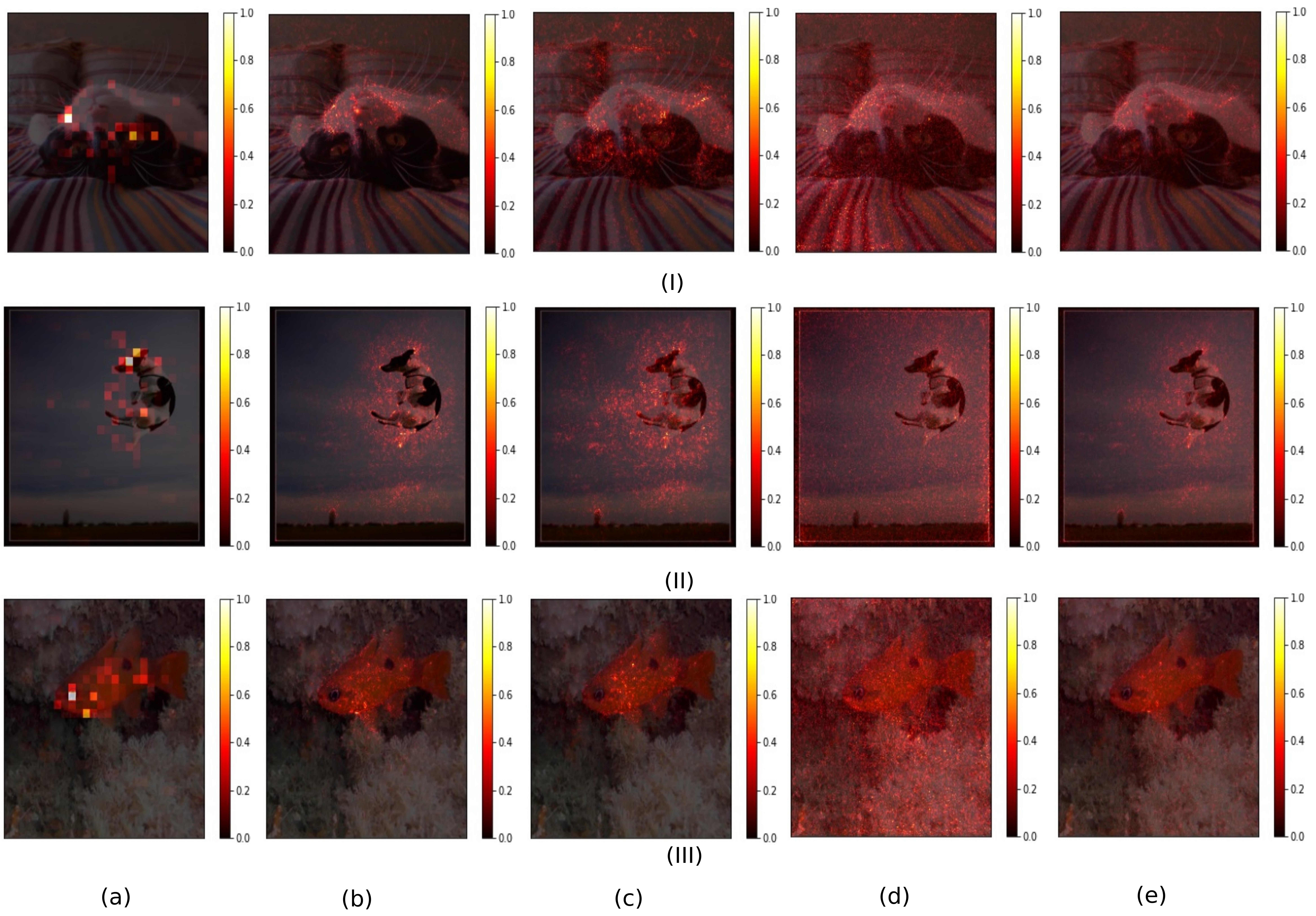}
    \caption{(I), (II), (III) are attribution maps overlaid on respective images of a Cat, a Dog and a Shark respectively: (a) SEA-NN map; (b) Integrated Gradient map; (c) Vanilla Gradient map; (d) Smooth Integrated Gradient map; (e) Agg-Mean map of (b)-(d) maps.}
    \label{ann-imgn}
\end{figure}

\begin{figure}[h!]
    \centering
    \includegraphics[scale= 0.15]{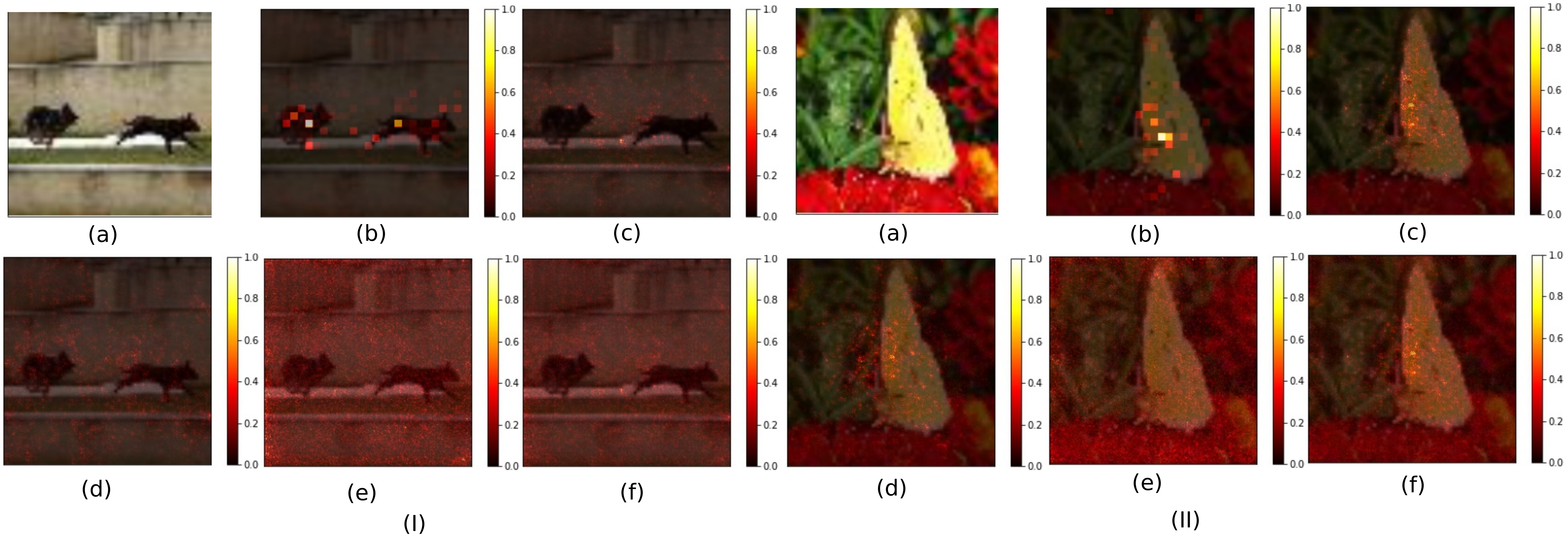}
    \caption{(I), (II) show images of German Shepherd dogs and a Sulphur Butterfly respectively with attribution maps overlaid on them: (a) Image; (b) SEA-NN map; (c) Integrated Gradient map; (d) Vanilla Gradient map; (e) Smooth Integrated Gradient map; (f) Agg-Mean map of (c)-(e) maps.}
    \label{dog-butter}
\end{figure}

In continuation to section 4 of the main paper, we present more qualitative results in this section. Figures \ref{golden-ret}, \ref{brain-flat}, \ref{ann-imgn} and \ref{dog-butter} show qualitative comparison of attribution maps for CUB, Brain Tumor Detection, Imagenet and Tiny Imagenet datasets respectively. Attribution maps of SEA-NN exhibit higher specificity than the attribution maps of other attribution methods.

\begin{figure}[h!]
    \centering
    \includegraphics[scale=0.25]{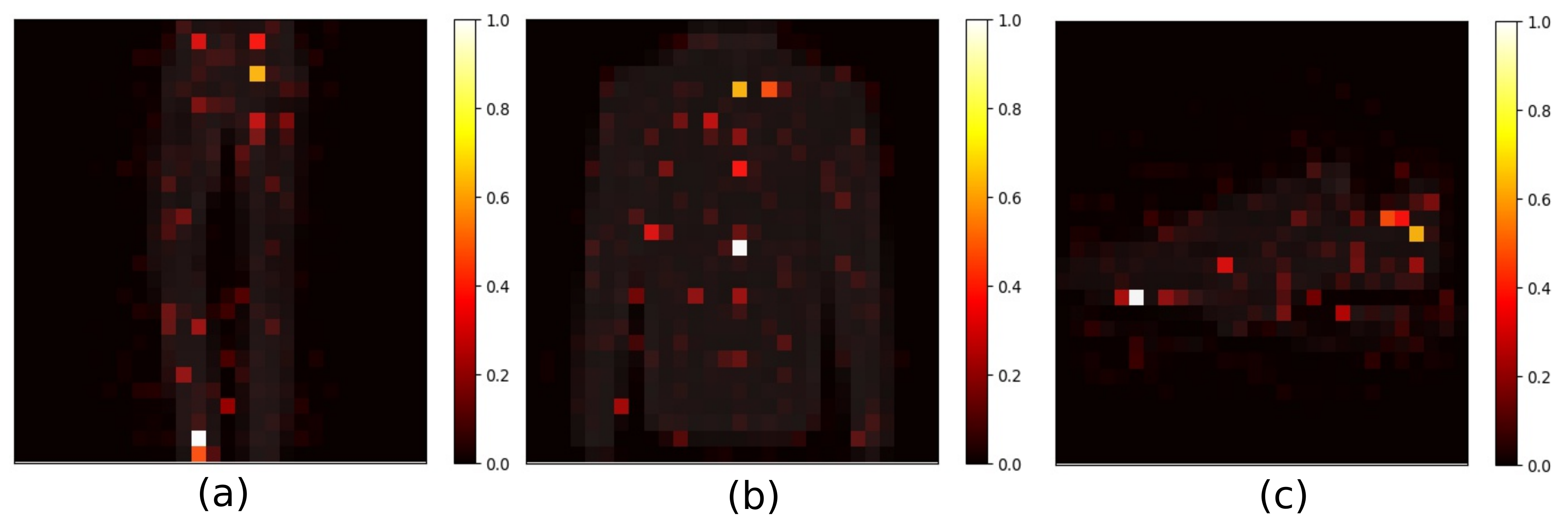}
    \caption{Global attribution maps of SEA-NN overlaid on images of corresponding classes (a) Trouser class (b) Coat class (c) Sneaker class}
    \label{global}
\end{figure}

\paragraph{Global Attribution using SEA-NN} In addition to providing local explanation for a specific input, our proposed attribution algorithm can also provide global explanation for an entire class in a dataset. Global explanations provide us with a more holistic understanding of the classifier NN. To extend SEA-NN as a global attribution algorithm, we only need to expand the training dataset for DSF to include heatmaps and their binary counterparts of all inputs belonging to a given class in the dataset. In Fig \ref{global}, we show global attribution maps of SEA-NN corresponding to three different FMNIST classes - Trouser, Coat and Sneaker; overlaid on randomly selected images belonging to these classes. The hyper-parameters for this experiment were the same as that of the FMNIST experiment included in section 4 of the main paper.

\begin{figure}[h!]
    \centering
    \includegraphics[scale = 0.2]{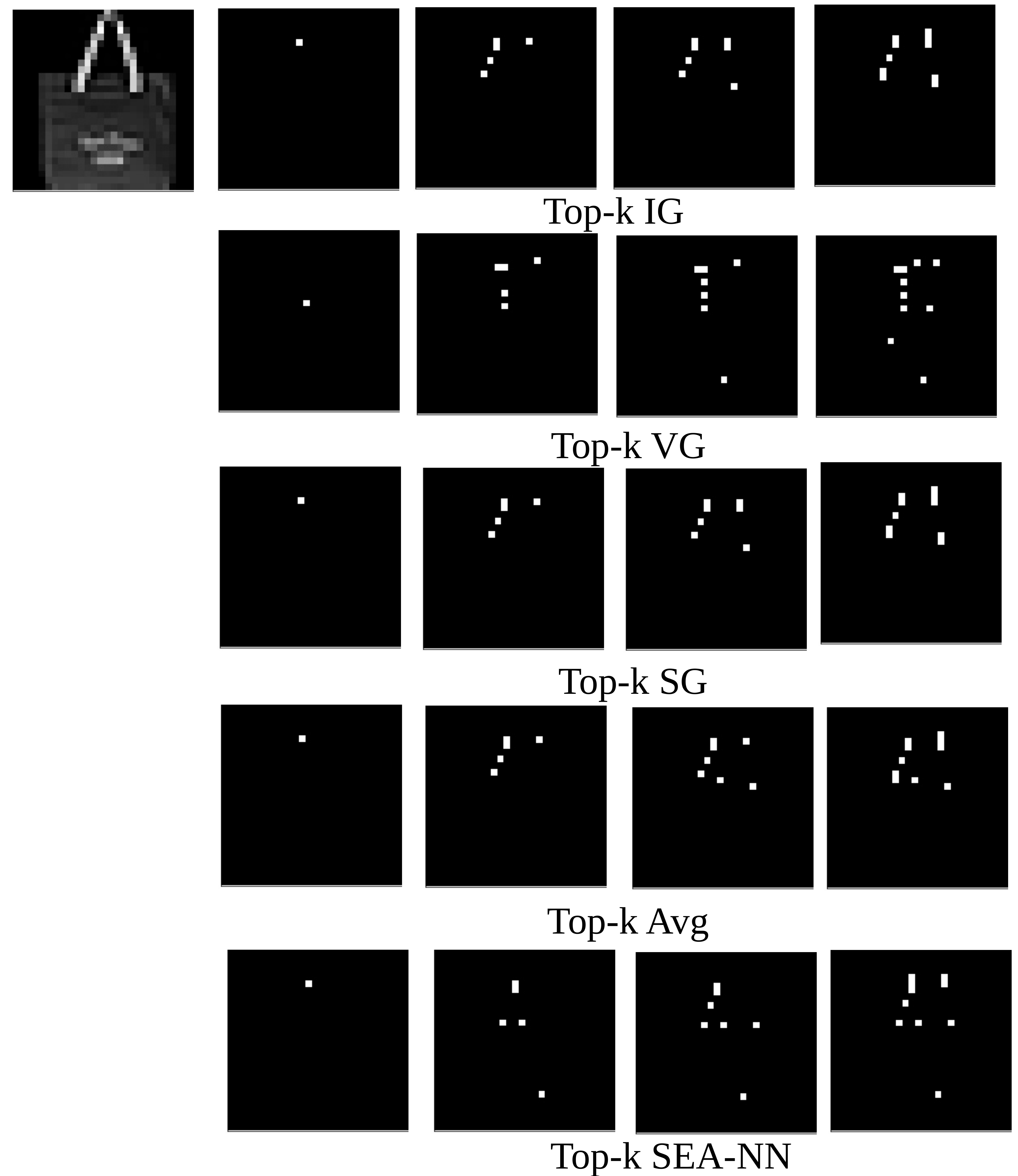}
    \vspace{-10pt}
    \caption{Top-$k$ pixels selected by attribution methods on image from Bag class in FMNIST. Columns from left to right represent $k$ as 1, 5, 7, 10 respectively}
    \vspace{-6pt}
    \label{top-k-supp}
\end{figure}

\paragraph{Visualizing Top-$k$ pixels}
In continuation to section 4 in the main paper, we present another result visualizing the top-$k$ pixels from FMNIST dataset. Fig \ref{top-k-supp} shows comparison of top-$k$ pixels for image of a Bag, as identified by different attribution maps. While most of the attribution maps only identified the handle of the bag, SEA-NN identifies a more diverse set of discriminatory pixels especially when $k$ is 10. The effect of feature-interaction and the benefits of submodularity help SEA-NN identify a more diverse set.

\subsection{More Quantitative Results}
We present some more details on the quantitative evaluation, some ablation studies and additional quantitative results in this section.

We tuned hyper-parameters of our method for a given dataset using a random subset of the same dataset. For computing average AUPC, we only considered the correctly classified images in case of FMNIST, CUB and Tiny Imagenet datasets. We considered all images for computing average AUPC on BTD as it was a relatively small dataset. In order to perturb pixels for computing AUPC, a popular approach is to choose top-$k$ pixels with the choice of $k$ as $\frac{1}{28}$th of the total number of pixels \cite{Petsiuk2018rise}. We followed this while computing AUPC for BTD. For datasets FMNIST, CUB, Tiny Imagenet, we perturb input regions of size $s~\textrm{x}~s$ following \cite{GohLWSB21Understanding} and consider total 8 such perturbations. The hyper-parameter $s$ was chosen as 16 for CUB and Tiny Imagenet. For FMNIST, $s$ was chosen as 8. 

\paragraph{Effect of Regularization Hyper-parameters}
We perform ablation studies on BTD to gauge the effect of regularization hyper-parameters $\lambda_1$ and $\lambda_2$ used for learning the scoring function (Eq. 2 in the main paper) and show results in Table \ref{ablation}. 

\paragraph{Comparing against more methods}
In Table \ref{aupc-more}, we show comparisons against more attribution methods in addition to the ones we compared against in Table \ref{aupc}.

\begin{table}[ht]
\caption{Area Under Perturbation Curve averaged over images of the datasets (lower is better)} 
\begin{center}
\begin{tabular}{lllll}
\textbf{Method} &\textbf{FMNIST} &\textbf{BTD} &\textbf{CUB} &\textbf{T.Img}\\
\hline
SEA-NN       & \textbf{3.46} & \textbf{11.11} & \textbf{4.82} & \textbf{5.42} \\
$input \circ gradient$          & 3.69 & 16.67 & 6.21 & 5.79\\
Deep-Lift         & 4.39 & 16.8 & 4.85 & 5.57\\
LIME          & 4.48 & 18.24 & 5.17 & 5.73\\
\hline
\end{tabular}
\end{center}
\label{aupc-more}
\end{table}

\paragraph{Robustness}
We conducted an experiment to test the robustness of attribution methods to noise in the input. We perturbed each input image of CUB by adding noise sampled from the uniform distribution, $\mathcal{U}(-0.02, ~0.02)$ and re-computed the attribution maps on these noisy inputs. For an attribution method that shows robustness to noise, we would expect these re-computed attribution maps to exhibit high similarity to the original attribution maps. To quantify this similarity, we computed Intersection over Union (IoU) scores between the thresholded versions of the re-computed attribution maps and the original attribution maps. Top 5000 pixels identified by the respective attribution maps were retained after thresholding these maps. For this experiment, we compare against Agg-Mean and Smooth Integrated Gradients (SG) as these methods have been observed to be relatively robust to noise \cite{Rieger2020ASD, smoothed-geometry-robust}. In Table \ref{rob}, we report the IoU scores averaged over the entire dataset. SEA-NN achieves the best average IoU score.

%

\begin{table}[ht]
    \centering
    \begin{tabular}{ll}\label{ablation}
    ($\lambda_1, \lambda_2$) & \textbf{AUPC}\\
    \hline
    (1, 1) & 11.91\\
    (1, 10) & 10.75\\
    (1, 100) & 11.47\\
    (10, 1) & 11.55\\
    (10, 10) & 11.43\\
    (10, 100) & 11.23\\
    (100, 1) & 11.46\\
    (100, 10) & 11.39\\
    (100, 100) & 11.15\\
    \hline
    \end{tabular}
    \caption{Ablation experiment on BTD}
    \label{ablation}
\end{table}

\begin{table}[ht]
\caption{Robustness comparison on CUB (higher average IoU is better)} 
\begin{center}
\begin{tabular}{ll}
\textbf{Method} & \textbf{Average IoU}\\
\hline
SEA-NN       & \textbf{2.94x$\mathbf{10^{-1}}$}\\
Agg-Mean      & 2.69x$10^{-1}$\\
SG          &  1.79x$10^{-1}$\\
\hline
\end{tabular}
\end{center}
\label{rob}
\end{table}

\paragraph{Additional Experiment}
We also got promising performance on evaluating our method on randomly chosen 100 images of the Benchmarking Attribution Methods (BAM) dataset \cite{BAM2019} using the Input Dependence Rate (IDR) metric proposed by them.
\begin{tabular}{c|c|c|c|c}
     \textbf{Metric} & \textbf{SEA-NN} & \textbf{Agg-Mean} & \textbf{IG} & \textbf{SG}  \\
     \hline
     IDR $\uparrow$ & \textbf{0.87} & 0.84 & 0.84 & 0.79
\end{tabular}

\begin{table}[h!]
\caption{Computation time (in sec) averaged across twenty randomly selected inputs.} \label{time}
\begin{center}
\begin{tabular}{lllll}
\textbf{Method} & \textbf{FMNIST (5-layer CNN)} & \textbf{BTD (VGG-11)} & \textbf{CUB (ResNet-18)} & \textbf{T.Img (ResNet-18)}\\
\hline
Proposed & 1.45x$10^{+1}$ & 1.48x$10^{+1}$ & 1.55x$10^{+1}$ & 1.56x$10^{+1}$\\
LIME & 3.89x$10^{-1}$ & 1.27x$10^{+1} $ & 1.26x$10^{+1}$  & 1.44x$10^{+1}$ \\
SG & 2.49x$10^{-2}$  & 6.50x$10^{-1} $ & 7.20x$10^{-1}$  & 7.28x$10^{-1}$ \\
IG & 1.11x$10^{-2}$ & 2.12x$10^{-1}$  & 9.64x$10^{-2}$   & 9.43x$10^{-2}$ \\
VG & 2.08x$10^{-3}$  & 4.03x$10^{-3}$  & 8.07x$10^{-3}$ & 8.35x$10^{-3}$\\
$inp\circ grad$      & 2.19x$10^{-3}$ & 3.99x$10^{-3} $ & 9.84x$10^{-3}$ & 9.61x$10^{-3}$ \\
GBP         & 2.17x$10^{-3}$  & 4.78x$10^{-3} $  & 9.39x$10^{-3}$  & 9.79x$10^{-3}$\\
Deconv          & 2.44x$10^{-3}$ & 5.05x$10^{-3} $ & 1.15x$10^{-2}$ & 1.17x$10^{-2}$\\
DL & 7.11x$10^{-3}$  & 1.84x$10^{-2} $ & 2.03x$10^{-2}$& 2.43x$10^{-2}$\\
\hline
\end{tabular}
\end{center}
\end{table}

\section{DETAILS OF COMPUTATION TIME}
In Table \ref{time}, we present a comparison of average computation time taken by different attribution methods on twenty randomly selected inputs from the datasets. These computation times were recorded after running experiments on an NVIDIA GTX 1080 Ti. As shown in Table \ref{time}, the computational time for SEA-NN becomes comparable to a popular attribution algorithm LIME for deeper classifiers. Furthermore, if we already have the down-sampled baseline attribution maps and their binary counterparts then the computation time for the proposed algorithm does not scale-up as the depth of the classifier NN increases. However, the computation time for other attribution methods increases as the classifier NN becomes deeper.

We note that attribution methods like XRAI \cite{xrai}, RISE \cite{Petsiuk2018rise} and SWAG \cite{swag} also require similar computational effort as the proposed algorithm. Our future work will include exploring a transfer learning approach so that a DSF trained on a subset of inputs can be used for the entire dataset. This will further reduce the computation time for us.

\end{document}